# Visual Analogy: Deep Learning Versus Compositional Models

**Nicholas Ichien**[*,1]
ichien@ucla.edu

**Qing Liu**[*,3]
qingliu@jhu.edu

**Shuhao Fu**[1]
fushuhao@ucla.edu

**Keith J. Holyoak**[1]
holyoak@lifesci.ucla.edu

**Alan Yuille**[3,4]
alan.yuille@jhu.edu

**Hongjing Lu**[1,2]
hongjing@ucla.edu

*Denotes equal author contribution
[1]Department of Psychology, [2]Department of Statistics
University of California, Los Angeles, Los Angeles, CA 90095 USA
[3]Department of Computer Science, [4]Department of Cognitive Science
Johns Hopkins University, Baltimore, MD 21218 USA

## Abstract

Is analogical reasoning a task that must be learned to solve from scratch by applying deep learning models to massive numbers of reasoning problems? Or are analogies solved by computing similarities between structured representations of analogs? We address this question by comparing human performance on visual analogies created using images of familiar three-dimensional objects (cars and their subregions) with the performance of alternative computational models. Human reasoners achieved above-chance accuracy for all problem types, but made more errors in several conditions (e.g., when relevant subregions were occluded). We compared human performance to that of two recent deep learning models (Siamese Network and Relation Network) directly trained to solve these analogy problems, as well as to that of a compositional model that assesses relational similarity between part-based representations. The compositional model based on part representations, but not the deep learning models, generated qualitative performance similar to that of human reasoners.

**Keywords:** analogy, visual reasoning, relations, computational modeling, deep learning

## Introduction

**Two Computational Approaches to Analogy** Analogical reasoning—the ability to recognize and exploit similarity based on the extrinsic *relations* that are shared between sets of entities, rather than the intrinsic features shared between individual entities—is widely viewed as a hallmark of human intelligence (Holyoak, 2012). To take a simple example, an arm and a body are analogous to a branch and a tree because an arm is an *extended part of* a body, just as a branch is an *extended part of* a tree. Computational models of analogy developed in cognitive science and artificial intelligence (AI) fall into two broad classes. One approach, popular in recent AI work, builds on deep neural networks that support training from raw input stimuli (e.g., image pixels, or words in a text) to a final task in an end-to-end manner. Learning in these networks is typically guided by minimizing errors in solving a particular task. This approach is now moving beyond tasks involving pattern recognition (such as object classification), for which deep learning has achieved great success, to reasoning tasks. The deep learning approach to analogy is to view it as a task for which a deep neural network can be trained end-to-end by providing massive datasets consisting of reasoning problems. This approach has been applied with some success to solving visual analogy problems, notably problems inspired by Raven's Progressive Matrices (RPM; Raven, 1938). After extensive training with RPM-like problems, deep neural networks have achieved human-level performance on test problems with similar basic structure (Santoro et al., 2017; Zhang et al., 2019; Hill et al., 2019). However, the success of these deep learning models depends on high similarity between training problems and test problems, and on datasets of massive numbers of RPM-like problems (e.g., 1.42 million problems in the PGM dataset, Barrett et al., 2018; and 70,000 problems in the RAVEN dataset, Zhang et al., 2019). For example, Zhang et al. (2019) used 21,000 training problems from the RAVEN dataset, and 300,000 from the PGM dataset.

This dependency on direct training in a reasoning task using big data makes the deep-learning approach fundamentally different from human analogical reasoning. When the RPM task is administered to a person, "training" is limited to general task instructions. Because the task is of interest as a measure of fluid intelligence—the ability to manipulate *novel* information in working memory—extensive pretraining on RPM problems is neither necessary nor desirable (Snow, Kyllonen, & Marshalek, 1984). Human analogical reasoning, more generally, is a prime example of zero-shot or few-shot learning—the ability to make inferences with minimal prior exposure to structurally similar problems.

An alternative approach to analogical reasoning is to view it not as a task on which to be trained directly, but rather as an inference problem based on computation of relational similarity. In this approach, the core of analogical reasoning is built upon compositional structures consisting of entities, their attributes, and relations between entities (e.g., Lovett & Forbus, 2017; Hummel & Holyoak, 1997). Analogy involves comparing structural representations of source and target analogs to assess their similarity. Although theoretically appealing, most models based on compositional structure are unable to extract relational representations directly from input images or texts, making it difficult to compare their performance to that of deep neural networks using the same training and test data.

To compare the two computational approaches, the present paper describes a new database of visual analogies based on pixel-level images of realistic 3D objects. This database allows generation of controlled analogy problems for human experiments, and also enables generation of a large amount of data for training deep neural networks. After obtaining benchmark data from a human experiment, analogy models based on deep learning are compared with a compositional model based on the extraction and comparison of whole-part relations.

**Visual Analogy with Objects and Object Parts** In previous studies of visual analogy, researchers have often employed simplified visual stimuli, such as combinations of geometric shapes in RPM-like problems (Carpenter, Just, Shell, 1990; Lovett & Forbus, 2017) and other meaningless two-dimensional forms (Bongard, 1970). The present study uses visual stimuli closer to realistic images of familiar three-dimensional objects—cars (see examples in Figure 1). These stimuli allow systematic manipulations of 3D object images, varying texture, shading, and viewpoint, while also enabling tight experimental control over the images used to construct analogy problems. At the same time, the stimulus set is extensive enough to provide a massive amount of data for training both deep learning models of analogy and a model based on comparison of compositional structures. All the computational models examined in the present paper avoid hand coding of representations, taking raw pixel-level images as inputs to solve analogy problems.

The analogy problems in this dataset focus on part-whole relations. In general, human perception and thinking show sensitivity to part-whole relations across both visual and semantic domains. Structural description theories of object recognition, which include part relations, provide a parsimonious explanation for viewpoint-invariant recognition in human vision (Biederman, 1987; Marr & Nishihara, 1973). Instances of basic level categories are unified by having a large number of shared parts. For example, Tversky and Hemenway (1984) found that people showed high agreement in rating the part goodness of meaningful car components, such as headlights and doors. Critically, part-whole relations also permit analogical reasoning. For example, children as young as four years old can map parts of a human body to their corresponding locations on trees and mountains (Gentner, 1977).

## Human Performance on Visual Analogy Task

### Methods

**Participants** 79 participants were recruited from Amazon Mechanical Turk ($M_{age}$ = 42, $SD_{age}$ = 11, age range = [24, 73], 41 female, 38 male). Participants provided online consent in accordance with the Institutional Review Board of the University of California, Los Angeles, and were compensated with a monetary reward.

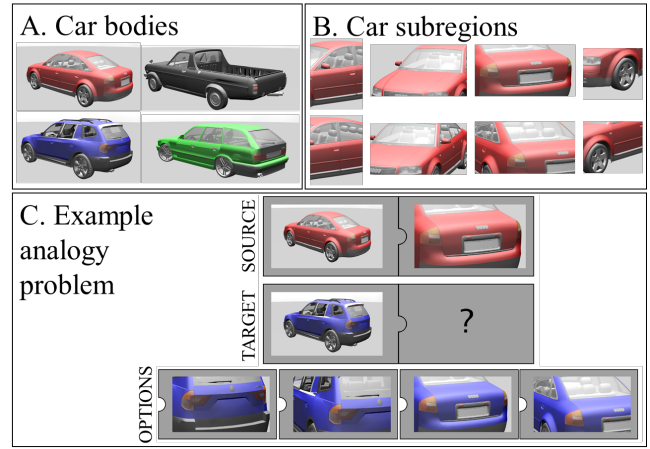

*Figure 1*. Experimental stimuli. Panel A: 3D car types including a sedan, truck, SUV, and station wagon. Panel B: subregions of the sedan car body. These subregions can be entire parts (top row), or pieces that do not correspond to entire parts (bottom row). Panel C: an example analogy problem in *A:B::C:?* format, constructed out of some of the images shown in panels A and B. The row of answer options includes the correct answer (leftmost), a wrong-subregion distractor (second), a wrong-car distractor (third), and a both-wrong distractor (rightmost).

**Materials and Procedure** Each problem was a four-term analogy in *A:B::C:?* format. The source analog (*A:B*) was an image of a car body paired with a subregion of it; the target was a different-style car body, paired with a set of four alternative images of subregions as options to complete the analogy. The problems systematically varied (1) the spatial alignment between images of source and target cars, (2) the visibility of analogous subregions in the whole-car images, and (3) whether or not images of analogous subregions depicted unitary car components.

Car images in analogy problems were generated using 3D car models taken from the ShapeNet 3D Core dataset (Chang et al., 2015). Figure 1A shows the four car types used in the experiment: sedan, SUV, truck, and station wagon. Figure 1B shows the different car subregions used to construct the analogy problems. Each subregion either fully (*parts* condition; top row) or partially (*pieces* condition; bottom row) depicted one of the following car components: door and window, hood and windshield, trunk and bumper, headlight and wheel.

Each analogy problem (Figure 1C) consisted of an incomplete 2 x 2 image array with the source car body in the top left corner, a subregion of that car body in the top right corner, the target car body in the bottom left corner, and a question mark in the bottom right corner. Each problem was based on one of four pairs of car bodies: sedan (source) and SUV (target; pictured in Figure 1C), SUV and wagon, wagon and truck, or truck and sedan. Below the array was a set of four answer options presented in a randomized order, and participants were asked to select the option that best fit the bottom right cell of the 2 x 2 array. In addition to the correct option, one option depicted a disanalogous subregion of the target car (*wrong subregion*), one option depicted an analogous subregion of the source car (*wrong car*), and a fourth option depicted a disanalogous subregion of the source car (*both wrong*). All options matched the target car in color. When the correct answer was a part subregion (e.g., top left image depicting a door and window in Figure 1B), the wrong-subregion distractor depicted the corresponding piece subregion (e.g., bottom left image depicting a partial view of a door and window in Figure 1B). These assignments were reversed when the correct answer was a piece subregion.

Our entire test set consisted of 128 analogy problems that varied three factors: orientation of source and target cars (same vs. different), subregion visibility in the corresponding whole car images (visible vs. invisible), and subregion type (part vs. piece). For visible problems, the source and target subregions were visible from the images respectively depicting the source and target car bodies, whereas for invisible problems these subregions were occluded in the corresponding images. For part problems, the source and target subregions fully depicted corresponding car components. Crossing each of these factors yielded 8 problem types (see examples in Figure 2). In order to avoid excessive fatigue, each participant was asked to complete 32 problems out of the possible 128. To that end, we created 4 32-problem subsets, each consisting of 4 problems falling into one of the 8 problem types. The 4 problem subsets were distributed evenly across participants and varied which of the 4 car body pairs were combined with which of the 4 car subregions on each of the 8 problem types.

Before starting the task, participants were given a practice analogy problem with line drawings of a gas pump and a lawnmower and a battery and a flashlight (instantiating the common relation *x powers y*). No training or practice on solving the car analogies was provided. No feedback was provided in the experiment, and we did not impose any time pressure on participants.

## Results

Participants achieved overall mean proportion correct of .61 ($SD$ = .21, range = [.09, .97], greatly exceeding the chance level of .25 correct. Figure 2 provides a breakdown of response selections for 8 experimental conditions. When participants did not select the correct answer, they more often selected one of the two distractors that included an element of the correct answer (correct subregion or else correct car body). They very seldom selected the option that was completely incorrect (wrong subregion of wrong car body). A three-way repeated measures ANOVA revealed three significant main effects, that accuracy was higher on same-orientation problems than on different-orientation problems, $F(1,75) = 34.20$, $p < .001$, on visible-component problems than on invisible-component problems, $F(1,75) = 7.95$, $p = .006$, and on part problems than on piece problems, $F(1,75) = 14.37$, $p < .001$. No interaction effects were significant.

## Model Performance on Visual Analogy Task

We implemented two deep learning models, a Siamese Network (Bromley et al., 1993) and a Relation Network (Santoro et al., 2017), which each instantiate an end-to-end approach to analogical reasoning based on extensive training with highly similar reasoning problems. We next implemented a Part-based Comparison Model (PCM), which is trained to segment objects into component parts, and then solves analogy problems by computing relational similarity between part-based feature vectors.

**Dataset for Training Deep Networks** We used the 3DComputerGraphicsPart dataset (Liu et al., 2021) to train the two deep learning networks. This dataset provides detailed part annotations for 3D CAD models of vehicles. Five subtypes of cars were selected from ShapeNetCore (Chang et al., 2015): sedan, SUV, wagon, truck, and minivan, each represented by a single car model. Each surface mesh on the CAD models was assigned a label from a set of 31 part segments (e.g., back left door, front right door, front left windows, right mirror, bumper). The images were rendered using Blender software using randomly selected textures. The virtual camera had a resolution of 1024×2048 and field of view of 90 degrees. The virtual environment showed one car at a time, with random background, lighting, camera position, and object texture/color.

To train the deep learning models, 30,000 analogy questions were created using part labels to generate images of subregions. We divided the set of 30,000 analogy questions into 27,000 for training and 3,000 for test. For each analogy question, we sampled two whole car images from different car types of the 3D CAD models (denoted as *A*, *C*). Then we randomly selected part or piece subregions of the car in the *A* image to generate the *B* image. The correct option for the *D* image was the unique image that yielded the same whole-part relation for the *C:D* image pair as for the *A:B* pair. For each analogy question, the other three alternative D images were randomly selected from 7 possible *D* images, including other part/piece images from the car in the *C* image, two subregion images from the car in the *A* image, and four subregion images from other car types (different from those in *A* and *C* images). None of the images used to train the networks were included in the visual analogy task to compare model and human performance.

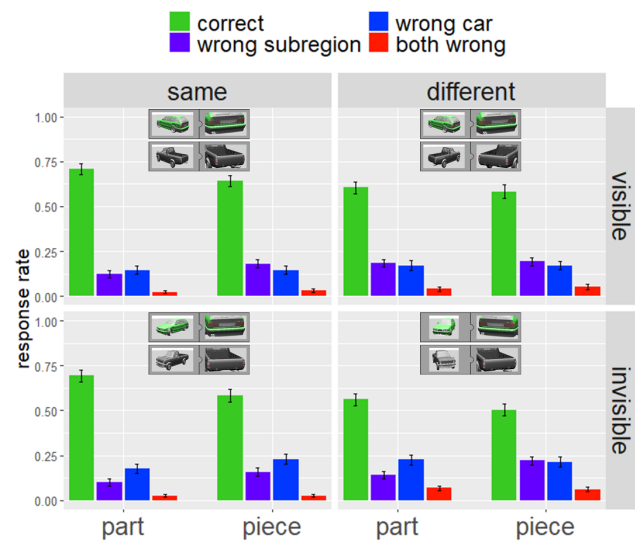

*Figure 2*. Human response selections for each problem type. Examples in each quadrant depict experimental condition. Error bars reflect ± 1 standard error of the mean.

**Siamese Network** A Siamese Network, which has been successfully applied to visual detection tasks (Bromley et al., 1993) and to visual and verbal analogy tasks (Sadeghi, Zitnick, & Farhadi, 2015; Rossiello et al., 2019), contains two or more identical subnetworks, emphasizing the role of comparisons among multiple inputs in forming comparable feature embeddings as visual representations. The embeddings are compared to assess the similarity between inputs. We adapted the model to learn to solve our car analogy problems. Figure 3A shows the architecture of the Siamese Network, which employs a VGG-16 network to translate pixel-level images to features. Features of whole-car images (*A* and *C* images) are processed by spatial pooling to form embeddings of size 4x4x512; subregion image features (*B* and *D*1-8 images) are pooled to form 1x1x512 embeddings. The feature embedding for image *B* is then expanded to the same size as that for image *A*, and then concatenated with *A* in the channel dimension. Feature embeddings for *D* images are similarly expanded and concatenated with the feature embedding for the C image. The concatenated features are passed through another convolutional layer and three fully-connected layers to obtain the final embedding of an image pair, which is a vector of size 512. We use contrastive loss with margin 1 to minimize the distance between concatenated feature embeddings from *A:B* images and embeddings from *C:D* images to choose the *D* image that best completes the analogy.

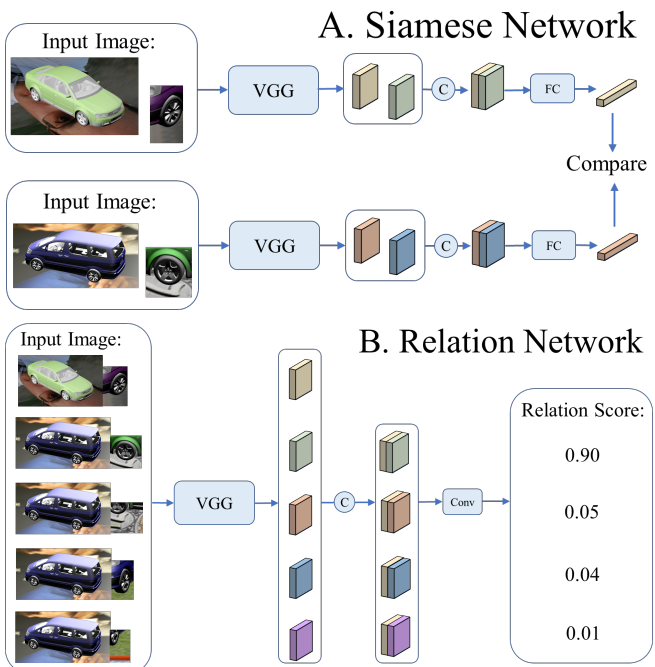

*Figure 3*. Siamese Network (Panel A), and Relation Network (Panel B) architectures for solving car analogy problems.

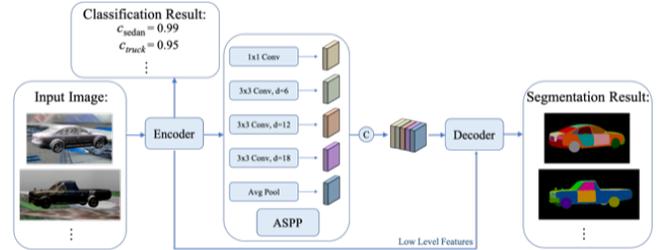

*Figure 4*. DeepLabv3+ Network Architecture for identification and segmentation of synthetic car images.

For both the Siamese Network, and the Relation Network described below, batch size was set to 4 analogy problems and the models were trained for 100 epochs. An Adam optimizer (Kingma & Ba, 2015) was used to learn network parameters, with $\beta_1 = 0.9$ and $\beta_2 = 0.999$. The learning rate started at 0.001 and reduced by a factor of 10 every 40 epochs. A color jitter augmentation was applied to all images before feeding them into the networks. Finally, we trained both networks end-to-end from scratch on the visual analogy training dataset.

**Relation Network** A Relation Network has been successfully applied to RPM-like analogy problems (Barrett et al., 2018) and query tasks (Santoro et al., 2017). We adopted the implementation by Sung et al. (2018) to set up the Relation Network to learn to solve the car analogy problems. The input to the Relation Network is the concatenation of features extracted from VGG-16 for the whole images (*A* and *C* images) and their corresponding subregions (*B* and *D* images). To train the network to solve analogy problems, we concatenated the whole-car image *C* with each of the candidate *D* images, and passed them through the network separately (see architecture shown in Figure 3B). The encoder extracted features from all image pairs, and then concatenated image features of the A:B image pair with the features of the C:D image pairs in the channel dimension. The concatenated features were then passed through additional convolutional layers and fully connected layers to estimate a relation score indicating the probability that a candidate *C:D* pair instantiates the same relation as the question pair of *A:B* images. We used cross entropy loss to train the Relation Network.

**Part-based Comparison Model**

To instantiate the structural comparison approach to analogy, we developed a part-based comparison model (PCM). This model employs image segmentation algorithms to extract visual features representing part-based structures for 3D objects, and then compares representations by computing a generic measure of relational similarity. From the same 3DComputerGraphicsPart dataset used to train the deep learning models, we rendered 40,000 synthetic images (30,000 for training and 10,000 for test; none were used in the analogy task) with automatically-generated part segmentation ground-truth. PCM is a variant of the DeepLabv3+ architecture, a deep neural network that is widely used for semantic segmentation in computer vision

(Chen et al., 2018). It includes encoder layers, an atrous spatial pyramid pooling (ASPP) module, and decoder layers (see Figure 4). An input image first is processed through an encoder module composed of a ResNet101 (He et al., 2016), yielding a feature map with 2048 channels and size (height and width) down-sampled by 16. Features extracted from ResNet101 are commonly used for image segmentation. The ASPP module then samples the input features at multiple spatial rates to gather information at different spatial scales. The outputs from different spatial scales are concatenated in the channel dimension and passed through the decoder layers. The output is a mask with the same height and width as the input image. Each pixel of the mask is assigned a label to indicate whether it belongs to the background or each of the 31 parts labeled in the dataset. We added a second output branch after the encoder layers of DeepLabv3+ to predict the car type label, which formed a regular object classifier.

We implemented PCM using Pytorch (Paszke et al., 2019) on two TitanX GPUs. Training was conducted using two standard cross-entropy losses: one for segmentation, and one for classification of car types. Batch size was set to 12 and the model was trained for 50 epochs. A stochastic gradient descent (SGD) optimizer was used to learn the network parameters, with momentum = 0.9 and weight decay = 0.0001. The learning rate started at 0.007 and decreased every epoch using a polynomial scheduler, power = 0.9. Before feeding the images into the network, standard data augmentation was applied (e.g., translation, scaling, cropping). We controlled the augmentation parameters to ensure the network only trained with whole car images (i.e., no partial car images were used during training). Car images used in the human experiment were excluded from the training set. The model achieved high performance for both part segmentation (mean intersect over union (mIoU) = 0.57) and subtype classification (accuracy = 0.99) on the test set.

We then applied the trained network to images used in the analogy problems to obtain segmentation and classification predictions. Example segmentation results are shown in Figure 5. When applied to the whole car images (images $A$ and $C$), the segmentations were reasonable, and the classification accuracy was 100%. When applied to the part/piece images (i.e., image $B$ and the alternative $D$ images), which provide incomplete visual input, the segmentations yielded small errors, and the classification accuracy dropped to 71%.

Next, we created compositional descriptions of images using the segmentation and classification results. We first converted the pixel-level labels in the resulting segmentation map to a low-dimensional feature vector. Specifically we counted the number of pixels that were parsed into each of the part segments, and computed the proportion for each part (i.e., number of pixels in a part divided by the total number of pixels in the resulting segmentation map). The dimensions of feature vectors were defined using a taxonomy with 13 parts for cars, adapted from a more generalized segmentation scheme used for parsing complex, real-world scenes in the PASCAL-Part dataset (Chen et al., 2014). The PASCAL segmentation aggregates over subsets of the 31 parts in the 3DComputerGraphicsPart dataset on which the part segmentation model was trained. For example, the PASCAL partonomy includes a door segment that aggregate over four segments, including back left door, back right door, front left door, and front right door. By concatenating the part-proportion vector with the car-type classification result, we obtained the 18-dimensional feature vector $f = cat(\mathbb{P}(m), c)$, where $m$ is the resulting segmentation map, $c$ is the car-type prediction, $\Sigma f_{1:13} = 1$ and $\Sigma f_{14:18} = 1$. Thus, for each analogy question, PCM represents images $I_A, I_B, I_C, I_{D1}, I_{D2}, I_{D3}, I_{D4}$ as feature vectors $f_A, f_B, f_C, f_{D1}, f_{D2}, f_{D3}, f_{D4}$, respectively.

Finally, to solve the visual analogy problem, a decision is derived by selecting the best $D$ image $\in \{D_1, D_2, D_3, D_4\}$ such that the relation that holds between image $A$ to image $B$ is similar to the relation between the two images $C$ to $D$. We computed the difference between feature vectors $f_A$ and $f_B$, and between $f_C$ and $f_D$, and used cosine distance to measure the dissimilarity of the two difference vectors. The same approach has been used in the Word2vec model (Zhila et al., 2013), and has proved effective in modeling visual analogical reasoning (Lu et al., 2019). The preferred answer $\widehat{D}$ is defined as the $D$ image that generates minimum cosine distance between difference vectors:

$$\widehat{D} = \underset{D \in \{D_1, D_2, D_3, D_4\}}{\arg\min} 1 - \cos(f_B - f_A, f_D - f_C)$$

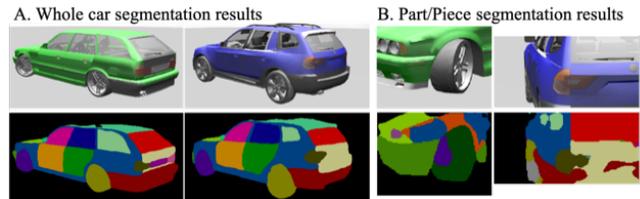

*Figure 5.* PCM segmentation results for images used in analogy problems. The model never saw these images during training.

### Analogy Performance for Models and Humans

All models (Siamese Network, Relation network, PCM) were tested on the same 128 visual analogy problems used in the human experiment, which involves car images never used in training for any of the models. Overall proportion correct was .38 for the Siamese Network, .53 for the Relation Network, and .61 for PCM. The latter model matched overall human accuracy (.61). More importantly, only PCM captures the qualitative differences in human accuracy among conditions. PCM, like humans, shows higher accuracy on same-orientation problems than on different-orientation problems, on visible-component problems than on invisible-component problems, and on part problems than on piece problems (see Figure 6). In order to quantitatively assess model fit to human data, we computed the root mean squared deviation (RMSD) between model accuracy and mean human accuracy on each of the 8 analogy problems types. This measure indicates how much each model deviates from human performance, with a lower RMSD indicating closer fit to human data. The

Siamese Network yielded an RMSD of .24 and the Relation Network .17, whereas PCM had an RMSD of .07, achieving by far the closest fit to human data.

To further evaluate the models as accounts of human analogical reasoning, we considered a possible non-analogical shortcut strategy for answering the problems. Using the problem in Figure 1C as an example, the problem could be answered correctly simply by selecting the *D* image most visually similar to the *B* image but not identical to it. We tested the possibility that any of the three models might have exploited this shortcut by providing only the subregion images (*B* image and *D* images) without showing the whole-car images (*A* and *C* images). The Siamese Network and PCM performed at chance on this test, indicating they did not exploit the shortcut. However, the performance of the Relation Network maintained the same (.53 correct) when the "analogy" problem was reduced to just two of the four terms. Thus, the Relation Network did not actually learn how to reason by analogy using relations at all.

## Discussion

PCM, a model based on segmenting object parts and identifying 3D objects from images, followed by comparison of part-based difference vectors, can account for the qualitative pattern of human accuracy on visual analogies between images of cars and their subregions. In contrast, two popular deep learning models that have been applied to visual analogies—a Siamese Network and a Relation Network— deviate seriously from human performance. Indeed, the Relation Network did not learn to reason by analogy at all, instead acquiring a non-analogical shortcut strategy. A methodological implication of these findings is that simply matching (or exceeding) human overall accuracy on some benchmark task is not a sufficient criterion for inferring that a computational model is simulating human cognitive mechanisms. It is important to compare model and human performance in greater detail at the level of controlled manipulations of problem types.

In contrast to PCM's approach to representation learning, a related approach is to explicitly learn representations of both object parts and the relations among them, using more computationally intensive but also more efficient learning processes such as analogical structure mapping (e.g., Chen, Rabkina, McClure, Forbus, 2019). By comparison, PCM relies on much simpler statistical learning processes to learn representations of object parts, offering a more parsimonious account of representation learning that does not depend on analogical reasoning to get off the ground. Importantly, the representations that PCM learned were expressive enough to provide the basis for a correspondingly simple but ultimately successful approach to analogy, in which we compared relations derived from those representations.

In PCM we built in this comparison procedure, implemented as a computation of cosine distance over difference vectors between two analogs. Another approach to solving analogies in zero-shot fashion, which requires less direct intervention from the modeler, is to employ meta-learning in which a model is trained both to solve and explicitly represent its solution to distinct but similar tasks, and then transfer that knowledge in order to solve novel tasks (e.g., Lampinen & McClelland, 2020).

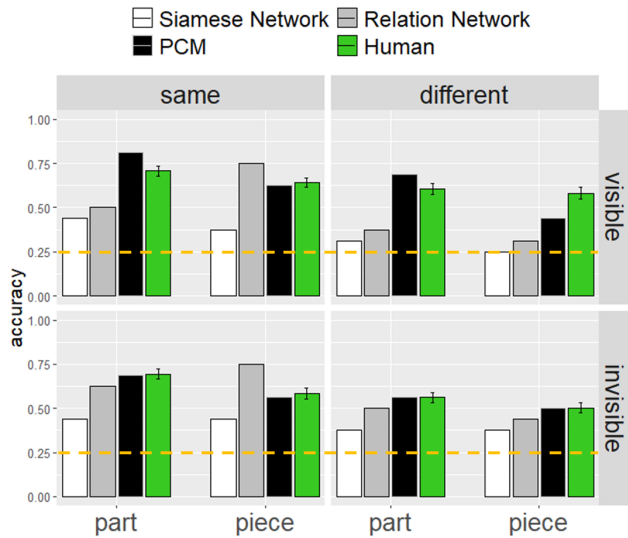

*Figure 6.* Model and human performance on the visual analogy task broken down by problem type. Dotted lines indicate chance performance. Error bars reflect ± 1 SEM for human data.

At any rate, the failures of the deep learning models of visual analogy examined here illustrate a general shortcoming of an approach that treats analogy as end-to-end learning of both perceptual stimuli representations and task structure. Instead of learning perceptual representations that might be generally useful in multiple tasks, these models acquire representations tailored to idiosyncrasies of the specific task used in training. The more promising approach, illustrated by PCM, is to first learn componential structure (here, part-whole relations for 3D objects) that are generally useful in distinctively different tasks (e.g., object recognition and segmentation). By training with varied tasks, the learned representations will acquire multi-task consistency. Although PCM does require extensive training to learn its compositional representations for visual inputs, we emphasize that it was not *trained* to solve analogies at all. Instead, PCM achieved superior performance on our visual analogy task simply by comparing its compositional representations using generic similarity measures. Generalizable analogical reasoning is unlikely to be achieved by deep learning with big data composed of a specific type of analogy problem. Rather, humans do (and machines might) achieve analogical reasoning by smart learning of representations that encode relational structure, coupled with efficient computation of relational similarity.

## Acknowledgments

Preparation of this paper was supported by NSF Grant BCS-1827374 to KH and HL, and NSF Grant BCS-1827427 to AL.